\documentclass{article}
\usepackage{spconf,amsmath,graphicx}
\usepackage{multirow}
\usepackage{diagbox}
\usepackage{booktabs}
\usepackage{url}
\usepackage{hyperref}

\title{Cross-lingual Alzheimer's Disease detection based on paralinguistic and pre-trained features}
\name{Xuchu Chen, Yu Pu, Jinpeng Li, and Wei-Qiang Zhang$^*$\thanks{* Corresponding author}\thanks{This work was supported by the National Key R\&D Program of
China under Grant No. 2020AAA0104500, and the National Natural Science Foundation of China under Grant No. 62276153.}}
\address{Department of Electronic Engineering, Tsinghua University, Beijing, China\\
\tt \small \{chen-xc20, puy19, lijp22\}@mails.tsinghua.edu.cn, wqzhang@tsinghua.edu.cn}
\begin{document}
\ninept
\maketitle
\begin{abstract}
We present our submission to the ICASSP-SPGC-2023 ADReSS-M Challenge Task, which aims to investigate which acoustic features can be generalized and transferred across languages for Alzheimer's Disease (AD) prediction. The challenge consists of two tasks: one is to classify the speech of AD patients and healthy individuals, and the other is to infer Mini Mental State Examination (MMSE) score based on speech only. The difficulty is mainly embodied in the mismatch of the dataset, in which the training set is in English while the test set is in Greek. We extract paralinguistic features using openSmile toolkit and acoustic features using XLSR-53. In addition, we extract linguistic features after transcribing the speech into text. These features are used as indicators for AD detection in our method. Our method achieves an accuracy of 69.6\% on the classification task and a root mean squared error (RMSE) of 4.788 on the regression task. The results show that our proposed method is expected to achieve automatic multilingual Alzheimer's Disease detection through spontaneous speech. 
\end{abstract}
\section{Introduction}
\label{sec:intro}

Alzheimer's Disease (AD) is a neurodegenerative disease. When suffering from AD, pathological changes occur in patients' brain, resulting in cognitive decline, expression degradation and other phenomena. Patients in different countries have similar symptoms. Clinical research shows that early treatment can delay the deterioration of AD. Therefore, the development of AD detection approach is crucial for the treatment of this disease. Although some researchers have carried out AD detection tasks on a single language \cite{ncmmsc2021git}, there are relatively few studies on cross-lingual AD detection. The academia still lacks a unified understanding of which speech features can be used for cross-lingual AD detection. The ICASSP-SPGC-2023 ADReSS-M challenge task aims to investigate how to extract generic acoustic features from speech for multilingual AD detection \cite{luz2023multilingual}.  

We focus on paralinguistic features and pre-trained features during the research. On the one hand, paralinguistic features have been proved reliable for distinguishing AD patients from healthy individuals \cite{chen2021automatic}. As a result of the disease, the patient's paralinguistic features have changed, and this change should be similar despite the patient's different language. On the other hand, monolingual pre-trained models such as Wav2vec2 and BERT have shown their effectiveness in monolingual AD detection \cite{pre2021adress}, and there are various multilingual pre-trained models that perform well in different downstream tasks. In addition, previous studies have confirmed that speeches of AD patients have linguistic features including fewer verbal real words, but more imaginary words, pauses and hesitations \cite{yuan2020disfluencies}. We therefore believe multilingual pre-trained models can extract effective pre-trained features from speeches and their transcribed texts. For this reason, we propose a cross-lingual AD detection method based on paralinguistic and pre-trained features.

The paper is organized as follows. The second section introduces our proposed methods. The third section shows the results. And the last section is a summary of this paper.

\section{methodology}
\label{sec:format}

In this section, we introduce our proposed AD detection methods based on paralinguistic and pre-trained features. The general flow diagram is shown in Fig. 1, which will be introduced in detail later. 

\begin{figure}[htb]

\begin{minipage}[b]{1.0\linewidth}
  \centering
  \centerline{\includegraphics[width=8.5cm]{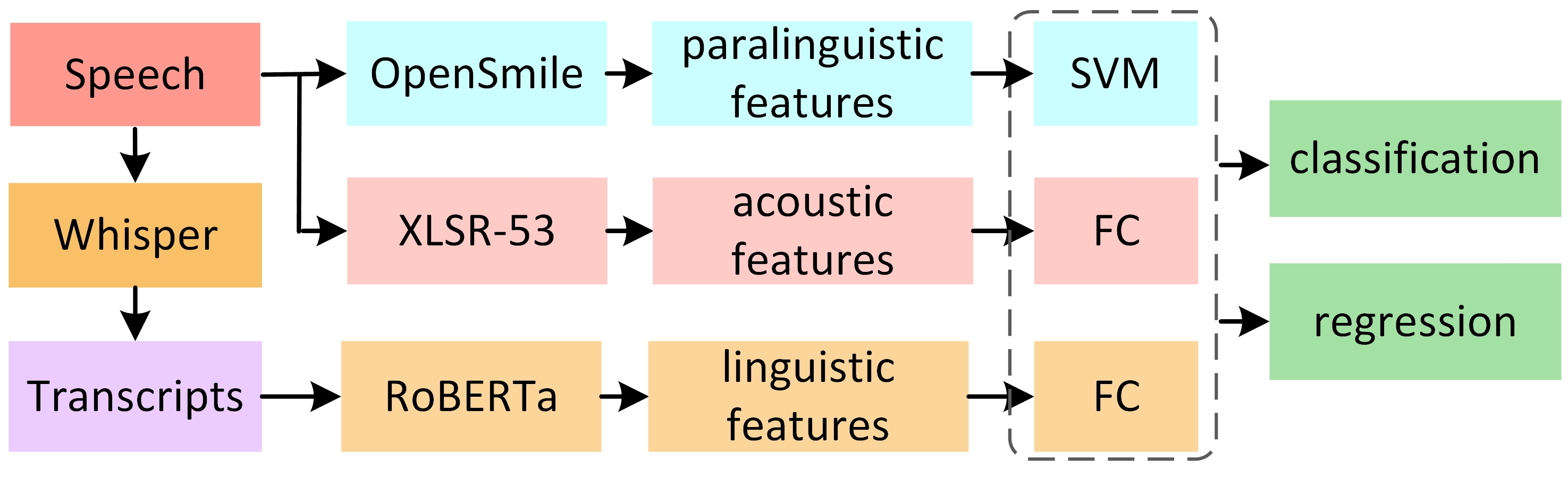}}
\end{minipage}
\caption{The general flow diagram of proposed methods. (FC: fully connected layer; SVM: support vector machine.)}
\label{fig:res}
\end{figure}


\subsection{Paralinguistic features based approach}
\label{sssec:subhead}

\begin{table*}[ht]
  \caption{Results of the classification task}
  \label{tab:result}
  \centering
  \begin{tabular}{ccccccc}
    \toprule 
    \textbf{No.} & \textbf{Methods}
    & \textbf{Accuracy}  & \textbf{Precision} & \textbf{Recall}  & \textbf{F1} \\
    \midrule
    1 & IS10-Paralinguistics-compat + SVM & $0.6957$ & $0.7273$ & $0.6667$ & $0.6957$ \\
    2 & IS10-Paralinguistics + SVM & $0.6957$ & $0.6923$ & $0.75$ &	$0.72$ \\
    3 & IS11-speaker-state + SVM & $0.6957$ & $0.75$ & $0.625$ & $0.6818$ \\
    4 & Whisper + RoBERTa + FC & $0.6522$ & $0.8333$ & $0.4167$ & $0.5556$ \\
    5 & XLSR-53 + FC & $0.5435$ & $0.5429$ & $0.7917$ & $0.6441$ \\

    \bottomrule
  \end{tabular}
  
\end{table*}

Research shows that paralinguistic features can be used as an important AD detection indicator for a single language \cite{chen2021automatic}. We use openSmile \cite{eyben2010opensmile} toolkit to extract paralinguistic features, which include: (a) IS10-Paralinguistics-compat feature set, (b) IS10-Paralinguistics feature set, and (c) IS11-speaker-state feature set. In Table 1, Method 1,2,3 use these three sets respectively. After feature extraction, we implement the standardization and employ support vector machine (SVM) as the classification and regression model.

\subsection{Pre-trained acoustic features  based approach}

At present, pre-trained models are widely used in speech recognition, affective computing and many other fields \cite{xlsr2022zhaojing}. We extract the pre-trained features using XLSR-53, a cross-lingual pre-trained model trained on speech data of 53 languages \cite{xlsr2020meta}. Its excellent performance on BABEL benchmark ensures its ability in cross-lingual feature extraction. Before input into the XLSR-53 model, the audio is cut into 6 second segments.  

 With the extracted features we train our downstream models, which contains two fully connected layers (FC). The classification and regression tasks are implemented by setting the number of neurons in the output layer to 2 and 1, respectively. For the classification task, we use the majority vote of all segments to predict the label of the original audio. For the regression task, we use the average MMSE score instead.

\subsection{Pre-trained linguistic features  based approach}
\label{ssec:subhead}

Whisper \cite{radford2022robust} is a speech recognition model and enables speech translation. We use the large-level Whisper model to transcribe the speech of the training set and the test set into English text.

RoBERTa \cite{liu2019roberta} is a pre-trained procedure for iterative BERT with longer training time and more data used for pre-training. Transcripts from the training set are used to fine-tune the pre-trained RoBERTa model, which is done by adding classification heads after the first embedding. The classification and regression tasks are accomplished by adjusting the number of neurons in the classification head.  


\section{Results}
\label{sec:pagestyle}

Table 1 shows the experimental result of each system on the classification task. By comparing the performance of each model in Table 1, it can be seen that the F1 score of Method 2 is the highest. 

Method 4 is to transcribe the speech into English text by Whisper, and then use RoBERTa for text classification. The precision of this method is high, but the recall is relatively low, which leads to a low F1 score. This may be resulted from the overfitting of the model due to the inconsistency between the pictures described by the subjects in the test set and the training set. And the performance of the whisper model can also have an impact on the results.

The accuracy of Method 5 is not so high, much inferior to the performance of pre-trained models on monolingual AD detection. The probable reason is that the training set of the XLSR-53 model did not contain Greek, and better results might have been obtained if the model was fine-tuned using Greek.

For the regression task, the best RMSE score in our method was 4.788, indicating that our method can be effectively used for cross-lingual Alzheimer's Disease detection.

In general, the paralinguistic features perform better than the pre-trained features in multilingual AD detection, probably because the low-level descriptors (LLDs) contained in the paralinguistic feature set can reflect the unexpected pauses in speeches regardless of languages. To the contrary, the high-dimensional embeddings extracted by pre-trained models have a stronger coupling with the language, which leads to a weaker generalization ability across languages. 

\section{Conclusion}
\label{sec:typestyle}

In this paper, we use paralinguistic and pre-trained features to detect Alzheimer's disease cross-lingually. Our proposed methods using paralinguistic features show excellent results, while the method based on pre-trained features does not show satisfactory results. The results suggest that paralinguistic features can be more effective for cross-lingual Alzheimer's Disease detection. Our future work includes the integration of the features mentioned above. And we will explore the use of ensemble models that focus on different features for cross-lingual Alzheimer's disease detection.

\bibliographystyle{IEEEbib}
\bibliography{refs}

\end{document}